\pdfoutput=1

\documentclass[dvipsnames,table]{article}

\usepackage{reportstyle}
\usepackage[utf8]{inputenc}
\usepackage[T1]{fontenc}

\usepackage{amsmath,amsfonts,bm}

\def\eqref#1{equation~\ref{#1}}

\def\1{\bm{1}}

\DeclareMathAlphabet{\mathsfit}{\encodingdefault}{\sfdefault}{m}{sl}
\SetMathAlphabet{\mathsfit}{bold}{\encodingdefault}{\sfdefault}{bx}{n}

\usepackage[labelfont=bf]{caption}
\usepackage{sectsty}
\usepackage{fancyhdr}
\let\cite\citep
\renewcommand{\headrulewidth}{1pt}
\makeatletter
\def\headrule{{\if@fancyplain\let\headrulewidth\plainheadrulewidth\fi
  \hrule\@height\headrulewidth\@width\headwidth\vskip-\headrulewidth}}
\makeatother

\definecolor{HYDarkBlue}{HTML}{2155EA}
\definecolor{HYLightBlue}{HTML}{A8DFF6}
\usepackage[
  colorlinks,
  linkcolor=HYDarkBlue,
  anchorcolor=HYDarkBlue,
  citecolor=HYDarkBlue,
  urlcolor=HYDarkBlue
]{hyperref}
\sectionfont{\color{HYDarkBlue}\fontfamily{zi4}\selectfont}
\subsectionfont{\color{HYDarkBlue}\fontfamily{zi4}\selectfont}
\subsubsectionfont{\color{HYDarkBlue}\fontfamily{zi4}\selectfont}
\usepackage{url}
\usepackage{booktabs}
\usepackage{multirow}
\usepackage{longtable}
\usepackage{tabularx}
\usepackage{enumitem}
\usepackage{graphicx}
\usepackage{float}
\usepackage{xcolor}
\usepackage{colortbl}
\usepackage{xspace}
\usepackage{pgfplots}
\usepackage{verbatim}
\usepackage{listings}
\usepackage{etoolbox}
\lstdefinestyle{promptbox}{
  basicstyle=\ttfamily\fontsize{10}{12}\selectfont,
  breaklines=true,
  breakatwhitespace=true,
  columns=fullflexible,
  keepspaces=true,
  showstringspaces=false,
  frame=single,
  framerule=0.4pt,
  framesep=3pt,
  xleftmargin=2pt,
  xrightmargin=2pt,
  aboveskip=6pt,
  belowskip=4pt
}
\makeatletter
\newcommand{\figcaptionhere}[2]{\par\nopagebreak{\def\@captype{figure}\caption{#1}\label{#2}}\par\medskip}
\makeatother
\usepackage{algorithm}
\usepackage[noend]{algpseudocode}
\pgfplotsset{compat=1.17}

\newcommand{\chk}[1]{#1}
\newcommand{\upd}[1]{#1}

\newcommand{\benchname}{InfraDev\xspace}

\providecommand{\probes}{hidden probes}

\providecommand{\gold}{golden patch}
\providecommand{\base}{pre-PR snapshot}
\providecommand{\post}{post-PR snapshot}

\title{\benchname: Can Coding Agents Develop AI Infrastructure, and How Can We Tell?}

\author{%
  \textbf{\small Yangshen Deng$^{1}$\thanks{Co-first authors.}, \quad Jiaqi Ruan$^{1}$\footnotemark[1], \quad
  Yeqi Huang$^{1}$\footnotemark[1], \quad Nick Lau$^{2}$, \quad Guangming Sheng$^{2}$, \quad Luo Mai$^{1}$}\\[0.45em]
  {\normalfont\small $^{1}$University of Edinburgh \quad $^{2}$Tencent Hunyuan}
  \vspace{0em}
}

\newcommand{\emailicon}{%
  \raisebox{-1.5pt}{\includegraphics[height=1.05em]{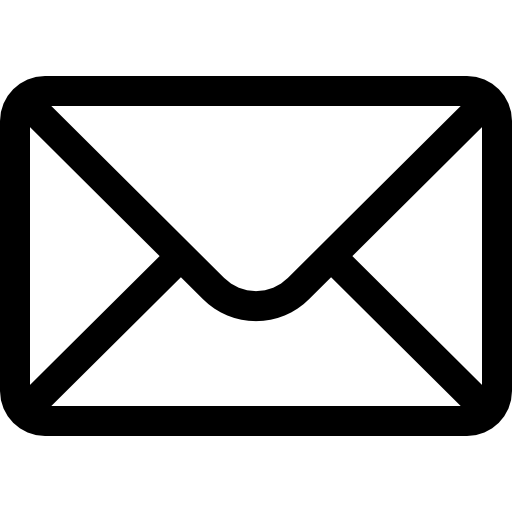}}}
\newcommand{\artifactlinks}{%
  \begin{tabular}{rl}
    \emailicon & \href{mailto:luo.mai@ed.ac.uk}{\nolinkurl{luo.mai@ed.ac.uk}},
      \href{mailto:petersheng@tencent.com}{\nolinkurl{petersheng@tencent.com}}
  \end{tabular}%
}

\begin{document}

\maketitle
\thispagestyle{fancy}
\vspace{-7pt}

\begin{abstract}
Coding agents are beginning to develop the production-grade AI
infrastructure that trains and serves large models, yet existing
benchmarks cover only single kernels or single-GPU tasks, and grade
with tests and prompts that misjudge both correct and incorrect
patches.
We introduce \benchname, a benchmark of 36 tasks built from real-world
pull requests to Megatron-LM, vLLM, and verl, covering infrastructure
development for training, inference, and reinforcement learning.
Each task asks an agent to reimplement a real change that spans many
files and must run correctly across many GPUs.
To grade such changes, we jointly synthesize hidden probes, which
enforce the requirements the original tests leave unchecked, and a
prompt that states every interface the probes call without revealing
the solution.
We also design a shared execution environment that reuses dependencies
and artifacts across tasks, cutting storage by \upd{52--57}\% and build time by \upd{13--14}\% on Megatron-LM and vLLM.
Across six frontier models, the best resolves only 38.9\% of the
tasks; the probes reject 32\% of the submissions that pass every
official test, and with the original pull request descriptions as
prompts, no model resolves a single task.
Failure analysis shows that structural and state errors dominate and
that agents' own tests rarely check memory, time, or traffic.
Turning these findings into an agent skill raises the resolved rate of
a frontier model on the Megatron-LM tasks by 10.5\%.
\end{abstract}

\section{Introduction}
\label{sec:intro}

Real-world AI infrastructure spans training, inference, and
reinforcement
learning~\citep{shoeybi2019megatron,kwon2023efficient,zheng2024sglang,sheng2025hybridflow}.
Each is backed by a large codebase that evolves rapidly with new models
and hardware: vLLM alone merged \upd{1{,}918} commits in June 2026, about \upd{64} per
day~\citep{luu2026vllm}. Developing this infrastructure means
implementing changes that range from a single GPU kernel to coordinated
edits across parallelism, memory, and communication subsystems. Such a
change must be correct in two senses. It must be \emph{numerically
correct}, computing the right activations, gradients, and states under
distributed execution, and it must be \emph{behaviorally correct}: a new
feature must take effect, and an optimization must actually reduce
memory, communication, or time on realistic workloads. Coding agents are
beginning to make these changes, as they take on longer and more complex
software engineering
tasks~\citep{jimenez2024swe,yang2024sweagent,wang2025openhands,swebenchpro,kwa2025measuring}.
This raises the question: \textbf{Can coding agents develop real-world
AI infrastructure, and how can we tell?}

Answering this question requires benchmarks that address three
limitations of existing evaluations.
First, existing benchmarks have a \textbf{narrow task scope}.
Most evaluate individual GPU
kernels~\citep{ouyang2025kernelbench,li2025tritonbench,chan2026parallelkernelbench,huang2026realistictritonbench}
or single-GPU inference
optimization~\citep{nangia2026isobench,yeon2026inferencebench}.
$\Phi$-Bench recently extended the scope to system-level
optimization~\citep{phibench2026}, but its tasks are optimization
problems with fixed targets rather than the changes developers
actually ship.
Most real infrastructure changes go further: they span many files and
modules and must run correctly across many GPUs.
Second, reusing a pull request's own tests and description as the
grading suite and task prompt~\citep{jimenez2024swe} yields
\textbf{unreliable grading} in both directions.
The tests are incomplete on both counts. Among the \chk{36} pull requests we
study, \upd{24} have no test that checks a numerical result, and
\upd{13} of the \upd{15} performance optimizations have no test that
checks the promised saving, so a patch that skips the optimization
still passes.
The prompts are underspecified. They omit interface names and
behaviors that the tests depend on, so a correct implementation can
fail.
The two problems are coupled, because every requirement a test
enforces must also be stated in the prompt.
Prior work strengthens tests~\citep{yu2025utboost,sweabs} or enriches
prompts~\citep{swebenchpro}, but not both together.
Third, evaluation on real infrastructure has \textbf{high cost}.
Each task pins its own dependencies and needs large artifacts such as
model weights.
Preparing them separately for every task takes substantial storage and time, which limits how many tasks a benchmark can afford and who can run it.

We address these limitations with three contributions, and then use
the resulting benchmark to understand why agents fail and to improve
them.

\textbf{A benchmark of large-scale AI infrastructure tasks.}
\benchname{} contains \chk{36} carefully selected tasks built from merged pull requests to
Megatron-LM, vLLM, and verl, covering training, inference, and
reinforcement learning. We start from the \upd{21{,}389} pull requests merged
into these repositories \upd{between January 2025 and September 2026}.
We keep those that change
core functionality, touch multiple files, add tests, and exceed \chk{1,000}
lines.
Each remaining task asks an agent to reimplement one pull request. A
task changes \upd{17.4} files and \upd{1{,}818} lines on average, and \upd{20} of
the \chk{36} require multiple GPUs to run.

\textbf{A method that builds tests and prompts together.}
We propose hidden-probe synthesis, implemented as an agent. For each
selected pull request, the agent first writes adversarial patches that
pass the official tests but violate requirements the tests leave
unchecked. It then writes new tests for these requirements, which we
call \textit{\probes}. A probe is kept only if it passes an acceptance
check: it must accept the golden patch in repeated runs and reject the
pre-PR snapshot, every adversarial patch, and a hollow patch that
defines the new interfaces without implementing them. Finally, the
agent writes a prompt that states every behavior the probes check and
every interface they call, without revealing how to implement them.
Both halves are necessary: the probes reject \upd{32}\% of the submissions
that pass all official tests, and with the original pull request
descriptions as prompts, no model resolves a single task.

\textbf{A resource-efficient execution environment.}
We design an execution environment that installs each task's dependencies into a virtual environment on a shared volume and allows tasks to share overlapping dependencies directly.
Model weights and datasets are likewise stored once in a shared cache and reused across tasks. This cuts storage by \upd{52--57}\% and environment build time by \upd{13--14}\% on Megatron-LM and vLLM, compared with building one image per task as prior benchmarks do~\citep{nangia2026isobench,yeon2026inferencebench}.

Using \benchname, we evaluate \chk{six} frontier coding agents. The best,
\upd{GPT-6 Sol, resolves} only \chk{38.9}\% of the tasks
(Section~\ref{sec:benchmark-result}). We then analyze the \chk{1{,}754}
failing test cases of the graded submissions and find failure modes
that agents repeat: structural errors are the most common, state errors
often escape the official tests, the tests agents write for themselves
rarely check memory, time, or traffic, and \chk{two thirds} of the errors are
shared by most models.

Finally, we encode these findings as \textbf{an agent skill for AI
infrastructure development}. The skill summarizes the failure cases and incorporate several development principles revealed by \benchname. Our experiments show that this skill raise the resolved rate of
GPT-5.6 Sol on the Megatron-LM tasks from 42.1\% to 52.6\%,
where the failure modes it targets are most common (Section~\ref{sec:skill}).

Encouraged by the above results, we will open-source \benchname{} and believe it offers a foundation for measuring and improving coding
agents on large-scale AI infrastructure.

\section{Related Work}
\label{sec:related}

\paragraph{Benchmarks for AI infrastructure and ML engineering.}
Most benchmarks in this area evaluate GPU kernel generation and
optimization~\citep{ouyang2025kernelbench,li2025tritonbench,xing2026flashinferbench,lange2025robustkbench},
extended to multiple GPUs by ParallelKernelBench~\citep{chan2026parallelkernelbench}
and to kernels inside their host frameworks by
RealisticTritonBench~\citep{huang2026realistictritonbench}.
ISO-Bench, InferenceBench, and $\Phi$-Bench move up to inference
optimization, but on a single GPU and with optimization targets fixed
in advance~\citep{nangia2026isobench,yeon2026inferencebench,phibench2026}.
A separate line evaluates agents on ML engineering, such as training
models for Kaggle competitions or reproducing
papers~\citep{mlebench2025,paperbench2025,rebench2025}; there the agent
uses infrastructure rather than changing it.
Despite similar names, InfraBench targets cloud and operations
agents~\citep{infrabench2026} and SWE-InfraBench targets
infrastructure-as-code~\citep{sweinfrabench2026}.
\benchname differs from all of these in its unit of work: a merged pull
request to a production training, inference, or RL framework, which
typically changes many files and must run correctly across GPUs.

\paragraph{Grading in repository-level benchmarks.}
SWE-bench and its descendants grade a patch with the tests that shipped
with the original pull request~\citep{jimenez2024swe,zan2025multiswebench,swebenchpro},
and infrastructure benchmarks extend them only with end-to-end checks of
accuracy and speed~\citep{nangia2026isobench,huang2026realistictritonbench}.
Both weaknesses of this recipe are known. On the test side, UTBoost adds
tests to raise coverage~\citep{yu2025utboost}, SWE-ABS uses adversarial
patches to show that many accepted patches are wrong~\citep{sweabs}, and
work on verifiable rewards reports policies that learn to pass weak tests
without solving the task~\citep{rewardhacking2025}. On the prompt side,
SWE-bench Verified removes underspecified tasks by human
annotation~\citep{swebenchverified} and SWE-Bench Pro adds requirement
and interface lists~\citep{swebenchpro}. Each fixes one side.
\benchname fixes both in one loop: adversarial patches find requirements
the official tests miss, hidden probes enforce them, and the prompt is
rewritten to state every interface the probes depend on
(\S\ref{sec:synthesis}). Environment reuse is a smaller but practical
concern: prior benchmarks build one image per
task~\citep{jimenez2024swe,nangia2026isobench} or share only a base
image~\citep{huang2026realistictritonbench}; we share dependency
environments and artifact caches across tasks (\S\ref{sec:env}).

\paragraph{Understanding and improving coding agents.}
Failure analyses of agents on SWE-bench and related benchmarks report
where runs break down, and agents are improved by better
interfaces~\citep{yang2024sweagent}, reusable skills and tool
libraries, or reflection over past attempts.
Kernel-generation systems apply the same ideas to one task
type~\citep{lange2025robustkbench,wang2025geak,baronio2025kevin}.
Our skill differs in where it comes from: it encodes the three failure
classes, contains no task-specific answers, and
targets the checks that agents skip, such as callers of a changed
function and the resource a change promises to save.

\section{\benchname}
\label{sec:design}

\begin{figure*}[!t]
 \centering
 \includegraphics[width=0.95\textwidth]{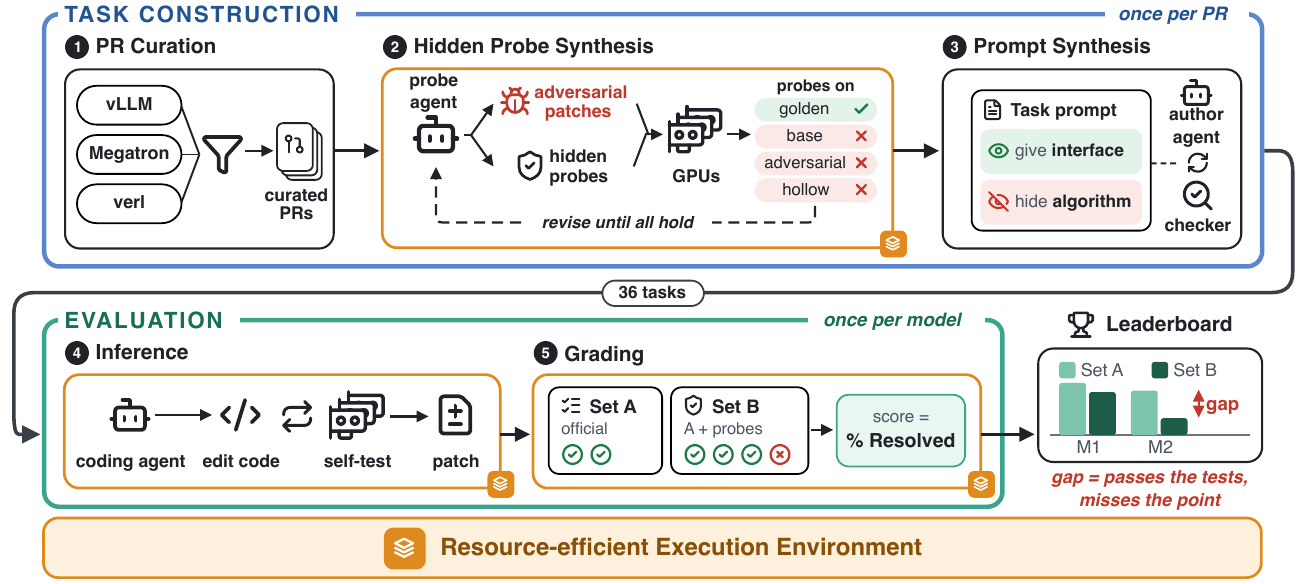}
 \caption{Overview of \benchname.}
 \label{fig:pipeline}
\end{figure*}

Each task in \benchname asks a coding agent to reimplement a real merged PR (pull request) from an AI infrastructure project.
Figure~\ref{fig:pipeline} shows how we build and evaluate these tasks.
We first curate challenging PRs that change core functionality of Megatron-LM, verl, and vLLM (\S\ref{sec:construction}).
For each PR, we jointly synthesize the tests and the prompt of the task.
An agent adds \probes{} that reject shortcut implementations of the PR.
Another agent writes a prompt that states what these tests check without revealing the solution (\S\ref{sec:synthesis}).
The agent under evaluation then implements the task in a multi-GPU sandbox, based on the prompt and the repository snapshot just before the PR.
We grade its patch with the \emph{official tests} of the PR and the \emph{\probes{}} (\S\ref{sec:evaluation}).
All stages run in a reusable execution environment (\S\ref{sec:env}) for storage saving and acceleration.

\subsection{Task Construction}
\label{sec:construction}

\subsubsection{PR Curation}

We collect tasks from three widely used production repositories that together cover the model life cycle: Megatron-LM for training~\citep{shoeybi2019megatron}, verl for reinforcement learning~\citep{sheng2025hybridflow}, and vLLM for inference~\citep{kwon2023efficient}.
We apply rule-based filters to all of their PRs and keep those that are merged and not reverted, change at least \chk{1{,}000} lines, and add new tests.
Guided by domain expertise, we then manually select the PRs that concern the core functionality of each framework.
This yields \upd{36} PRs, most of which add new features or optimize performance (Table~\ref{tab:pr-types}).
For each PR, we record the \base{} just before the merge and the \post{} right after it.
Their difference is the \emph{\gold{}}, the change the developers actually made, which serves as the reference solution.

\begin{table}[t]
\centering
\small
\caption{Task types of the \upd{36} selected pull requests. \emph{New feature}: adds
functionality that the \base{} lacks; \emph{Performance}: keeps the functional result but
reduces time, memory, or communication; \emph{Refactor}: restructures code without adding
functionality; \emph{Bug fix}: corrects wrong behavior.}
\label{tab:pr-types}
\resizebox{0.8\linewidth}{!}{
\begin{tabular}{@{}llllll@{}}
\toprule
\multicolumn{1}{@{}c}{Repository} & \multicolumn{1}{c}{New feature} & \multicolumn{1}{c}{Performance} & \multicolumn{1}{c}{Refactor} & \multicolumn{1}{c}{Bug fix} & \multicolumn{1}{c@{}}{Total} \\
\midrule
Megatron-LM (training) & 9  & \upd{8}  & 1 & 1 & \upd{19} \\
verl (post-training)   & 5  & 2  & 0 & 0 & \chk{7}  \\
vLLM (inference)       & \upd{5}  & 5  & 0 & 0 & \upd{10} \\
\midrule
Total                  & \upd{19} & \upd{15} & 1 & 1 & \upd{36} \\
\bottomrule
\end{tabular}
}
\end{table}

\subsubsection{Joint Synthesis of Tests and Prompt}
\label{sec:synthesis}

\textbf{Hidden-probe synthesis.}
We judge whether a modification completes a task by two criteria.
\textit{Behavioral correctness} means that the program behaves as the task requires when it runs: a new feature must take effect, and an optimization must actually reduce memory, communication, or time.
\textit{Numerical correctness} means that the program computes the right values, such as activations, gradients, and model parameters.

The official tests of the selected PRs are weak in both criteria.
First, numerical checks are often absent: in 24 of the \upd{36} evaluated tasks, no official assertion verifies the numerical result of the modification.
Second, behavioral checks are rare: in \upd{13} of the \upd{15} PRs that optimize performance, the official tests never check whether the optimization saves time, memory, or communication.
An implementation that computes the right output but omits the optimization therefore still passes them.

We address this gap with an agent skill that iteratively synthesizes \emph{\probes{}}, tests for requirements that a task imposes but its official tests leave unchecked (Algorithm~\ref{alg:probe-synthesis}).
The agent first summarizes what the PR promises to implement (line~\chk{1}).
It then builds \emph{adversarial patches}, incorrect implementations that pass every official test but violate numerical or behavioral correctness (line~\chk{2}).
Next, the agent writes probes that check both criteria (line~\chk{3}).
Appendix~\ref{app:probes} describes how a probe decides whether a result is correct.
For example, verl PR~6974 syncs weights from the trainer to the rollout engine by sending only the entries that changed since the last step.
Its probe counts the sent entries and rejects an adversarial patch that sends \chk{65{,}536} entries when only \chk{655} have changed.

We then refine the probes and the adversarial patches together (lines~\chk{4--5}).
In each round, the skill runs a check, and the agent revises every probe or adversarial patch that fails.
The check requires each probe to pass on the \gold{} and fail on every adversarial patch.
It also requires each adversarial patch to break a promise of the PR while still passing the official tests.
Appendix~\ref{app:probes} lists the full check.

\begin{algorithm}[t]
\caption{Hidden-probe synthesis. }
\label{alg:probe-synthesis}
\begin{algorithmic}[1]
\Require PR $r$, its \gold{} $\Delta_{\mathrm{gold}}$, the \base{} $\mathcal{S}$, and the official tests $T_{\mathrm{off}}$
\Ensure \probes{} $T_{\mathrm{probe}}$
\State $C \gets \textsc{Summarize}(r)$ \Comment{what $r$ promises to implement}
\State $\Delta_1, \dots, \Delta_k \gets \textsc{BuildAdversarialPatches}(C, T_{\mathrm{off}})$ \Comment{wrong patches that pass $T_{\mathrm{off}}$}
\State $T_{\mathrm{probe}} \gets \textsc{WriteProbes}(C)$ \Comment{check both criteria}
\While{not $\textsc{CheckProbes}(T_{\mathrm{probe}}, \Delta_1, \dots, \Delta_k)$}
  \State Revise $T_{\mathrm{probe}}$ and $\Delta_1, \dots, \Delta_k$ to fix the failed checks
\EndWhile
\State \Return $T_{\mathrm{probe}}$
\end{algorithmic}
\end{algorithm}

\textbf{Prompt synthesis.}
To improve the quality of prompts, we rewrite each PR as a task prompt, the agent's only description of the change.
The prompt must be \emph{sufficient}: a correct implementation must not fail because the prompt omits an interface or requirement on which the tests depend, such as the name of a function they import.
It must also be \emph{solution-free}: it must not explain how to meet the requirements, because such guidance does part of the task and makes weaker agents appear capable.
We meet both requirements by splitting the information about a PR into a \emph{specification} and a \emph{solution} and putting only the specification in the prompt.
The specification states \emph{what} the change must do: the interfaces the tests call, the behaviors the official tests and \probes{} check, and the execution rules, such as the definition of done.
The solution describes \emph{how} to do it: algorithms, formulas, numerical constants and tolerances, and validation strategies.
We omit any item that could belong to either.

A second agent skill constructs the prompt.
It receives from probe synthesis a handoff record that lists the claimed behaviors and the required symbols with input types, but not the probe code.
A static analyzer first extracts the \emph{interface contract}: the symbols that the \gold{} introduces and that the official tests or the \probes{} reference.
An authoring agent then writes the prompt from the PR description, the \gold{}, the official tests, the handoff record, and the contract.
It states each claimed behavior either as a requirement or as a decision explicitly left to the agent.
Finally, the prompt must pass two checks: a linter rejects any prompt that omits a contract symbol, and a checker agent rejects any prompt that discloses part of the solution.
The author revises the prompt until it passes both checks.
Appendix~\ref{app:task-prompt} gives the unified task-prompt template filled by this procedure.

\subsection{Evaluation}
\label{sec:evaluation}

\textbf{Inference and Grading.} We prepare Docker containers for all agents and set a \chk{2.5} hours time limit for each PR. Recovery mechanism is also introduced to prevent agent stopped accidentally in early stage. Agent's output patch is applied to the \base{} in a separate container.
The test directory is then replaced by that of the \post{} for evaluation, so agents cannot cheat by modifying test.
To quantitatively evaluate how good agents solve the task, grading will evaluate patches with real GPUs.

\textbf{Metrics.}
We grade each patch on the official tests together with the \probes{}.
For a task $p$, let $N_p$ be the number of test cases collected when the \gold{} runs, including the cases that the \gold{} itself fails or skips on our hardware.
The pass rate of a patch is $s_p = \mathrm{passed}_p / N_p$, and \% Passed averages $s_p$ equally over tasks.
A task is resolved when the patch passes at least as many test cases as the \gold{}, so the patch does not have to pass the cases that the \gold{} fails.
\% Resolved is the share of resolved tasks.
A missing, empty, or inapplicable patch scores $0/N_p$ instead of being dropped, since dropping it would reward models that give up.
Harness faults are excluded and counted separately, so an infrastructure error is never recorded as a model failure.
We keep the per-test outcomes and the full agent trace of every run for the failure analysis in \S\ref{sec:current-cases}.

\subsection{Execution Environment}
\label{sec:env}

\textbf{Limitation of Docker.}
Repository-level benchmarks usually build one Docker image per task.
Docker caches image builds as a prefix, reusing the build steps before the first changed step and rerunning every step after it.
This prefix cache works poorly for AI infrastructure projects, whose dependencies change often.
For example, the \upd{19} Megatron-LM tasks were merged within \chk{five} months but use \chk{six} versions of Transformer Engine.
Each version change makes Docker rerun dependency steps after it.
Moreover, tests of production infrastructure need large artifacts, such as model weights and datasets.
The test code downloads these artifacts while it runs, so Docker does not see them and cannot cache them across tasks.

\textbf{Resource-efficient Execution Environment.}
We break the prefix reuse of Docker and reuse each package and each artifact independently, so a change to one of them never forces the others to be rebuilt or stored again.
We maintain several lightweight base Docker images for each AI infrastructure repository, containing only components shared across multiple PRs, such as CUDA and the Python interpreter.
The dependencies of each task go into a virtual environment that \texttt{uv} installs on a volume shared by all containers.
Tasks with the same hash of their locked dependencies share one environment, and a new environment hard-links cached packages instead of reinstalling them.
Every container also mounts one shared Hugging Face cache, so each model or dataset is downloaded only once.
Both agent inference and grading build their environments this way.

\section{Experiment}
\subsection{Setup}

\textbf{Models and harnesses.}
We evaluate \chk{six} coding agents on \benchname{}, \chk{two} built on frontier closed-source models and \chk{four} on open-source models.
GPT-6 Sol and GPT-5.6 Sol run in OpenAI's Codex harness; GLM~5.3, DeepSeek V4 Pro (0813), Kimi~K3, and HY4 preview run in the OpenCode harness.\footnote{We use Codex version \chk{0.155.1} and OpenCode version \chk{1.18.31}.}
We set the reasoning effort of every model to \texttt{xhigh} or a comparable level.
Each agent runs once on each task, and we call the patch it produces a \emph{submission}.

\textbf{Environment.}
\upd{All experiments run on nodes with eight NVIDIA H800 or H20 GPUs each.}
Each inference or grading run has exclusive use of one node.
An agent has \chk{2.5} hours per task, excluding environment preparation; a run that reaches this limit is graded on the patch it has produced.
A task's golden calibration and the grading of its submissions use the same environment and GPU type.

\subsection{Results on \benchname}
\label{sec:benchmark-result}

\begin{table}[t]
\centering
\setlength{\aboverulesep}{0pt}
\setlength{\belowrulesep}{0pt}
\renewcommand{\arraystretch}{1.3}
\caption{Results of \chk{6} models on \benchname{}. \% Resolved is the share of tasks on which every test passes; \% Passed is the share of tests passed on a task, averaged over tasks. Time, Cost, and Tokens are per-task means of the agent's run time, API cost estimated at list prices, and tokens consumed.
}
\label{tab:resolution}
\resizebox{0.75\linewidth}{!}{
\begin{tabular}{@{}l>{\columncolor{black!8}}lllll@{}}
\toprule
\multicolumn{1}{@{}c}{Model} & \multicolumn{1}{>{\columncolor{black!8}}c}{\% Resolved} & \multicolumn{1}{c}{\% Passed} & \multicolumn{1}{c}{Time (min)} & \multicolumn{1}{c}{Cost (\$)} & \multicolumn{1}{c@{}}{Tokens (M)} \\
\midrule
GPT-6 Sol & \underline{\textbf{\chk{38.9}}} & \textbf{\upd{78.8}} & \underline{\textbf{\chk{43.9}}} & \chk{5.37} & \underline{\textbf{\chk{20.86}}} \\
GPT-5.6 Sol\footnotemark & \chk{30.6} & \upd{78.6} & \textbf{\chk{73.9}} & \chk{17.98} & \textbf{\chk{24.71}} \\
GLM 5.3 & \chk{25.0} & \chk{76.7} & \chk{136.9} & \chk{13.89} & \chk{48.84} \\
HY4 preview & \textbf{\chk{36.1}} & \underline{\textbf{\chk{83.3}}} & \chk{118.1} & \textbf{\chk{2.09}} & \chk{31.75} \\
DeepSeek V4 Pro & \chk{30.6} & \chk{74.3} & \chk{94.1} & \underline{\textbf{\chk{1.89}}} & \chk{41.80} \\
Kimi K3 & \chk{30.6} & \upd{75.8} & \upd{136.4} & \upd{13.48} & \upd{33.58} \\
\bottomrule
\end{tabular}
}
\end{table}

\footnotetext{GPT-5.6 Sol was measured through the Codex coding plan between September \chk{21} and \chk{24}, \chk{2026}; we observed possible degradation of the served model around these dates.}
\begin{table}[t]
\centering
\small
\setlength{\tabcolsep}{6pt}
\caption{\% Resolved per task category. The number of tasks is in parentheses.}
\label{tab:resolution-by-category}
\begin{tabular}{@{}llll@{}}
\toprule
\multicolumn{1}{@{}c}{Model} & \multicolumn{1}{c}{Training (\chk{19})} & \multicolumn{1}{c}{Inference (\chk{10})} & \multicolumn{1}{c@{}}{RL (\chk{7})} \\
\midrule
GPT-6 Sol & \underline{\textbf{\chk{52.6}}} & \chk{10.0} & \textbf{\chk{42.9}} \\
GPT-5.6 Sol & \textbf{\chk{42.1}} & \chk{10.0} & \chk{28.6} \\
GLM 5.3 & \chk{26.3} & \chk{10.0} & \textbf{\chk{42.9}} \\
HY4 preview & \textbf{\chk{42.1}} & \chk{10.0} & \underline{\textbf{\chk{57.1}}} \\
DeepSeek V4 Pro & \textbf{\chk{42.1}} & \chk{10.0} & \chk{28.6} \\
Kimi K3 & \chk{31.6} & \chk{10.0} & \underline{\textbf{\chk{57.1}}} \\
\bottomrule
\end{tabular}
\end{table}

 Table~\ref{tab:resolution} shows that \benchname{} remains challenging for all models: the resolved rate ranges from \chk{25.0}\% for GLM~5.3 to \chk{38.9}\% \upd{for GPT-6 Sol}.
Across task categories (Table~\ref{tab:resolution-by-category}), inference tasks are the hardest: \upd{every model resolves only one} of the \chk{ten}.
Resolved rates spread more widely on training tasks (\chk{26.3}\% to \chk{52.6}\%) and on RL tasks (\upd{28.6}\% to \chk{57.1}\%).
No single model leads in all categories: GPT-6 Sol performs best on training tasks, \upd{and HY4 preview and Kimi~K3 tie for best on RL tasks}.

A high passed rate does not mean that a model completes tasks.
Every model passes between \upd{74.3}\% and \upd{83.3}\% of the tests, but none resolves more than \chk{38.9}\% of the tasks, because a task is resolved only when all its tests pass.
\upd{Nearly equal} passed rates can also hide different resolved rates: GPT-6 Sol and GPT-5.6 Sol \upd{pass} \upd{78.8}\% and \chk{78.6}\% of the tests but resolve \chk{38.9}\% and \chk{30.6}\% of the tasks.
The models differ on a few precise tests, which a patch that only looks complete cannot pass.

No model we test leads in resolved rate, time, and cost at once.
GPT-6 Sol comes closest: it \upd{has} the highest resolved rate, is the fastest at \chk{43.9} minutes per task, and costs less than GPT-5.6 Sol, GLM~5.3, and Kimi~K3.
HY4 preview \upd{resolves one task fewer} at \upd{39}\% of the cost but takes \chk{2.7}$\times$ as long.
Cost and time also depend on the provider's prices, caching, and serving speed, not only on the model.

\subsection{Ablation Study}
\label{sec:ablation}

\begin{figure}[t]
\centering
\begin{minipage}[b]{0.51\linewidth}
\centering
\begin{tikzpicture}[baseline=(current axis.south)]
\begin{axis}[ybar=0.5pt, scale only axis, width=0.86\linewidth, height=2.5cm, ymin=0, ymax=90, ytick={0,20,...,80}, bar width=6pt,
  enlarge x limits=0.07, area legend, ymajorgrids, grid style={dashed,gray!30},
  symbolic x coords={gpt6,gpt,glm,hy4,ds,kimi}, xtick=data,
  xticklabels={{GPT-6\\Sol}, {GPT-5.6\\Sol}, {GLM\\5.3}, {HY4\\preview}, {DeepSeek\\V4 Pro}, {Kimi\\K3}},
  x tick label style={font=\footnotesize, align=center, inner sep=1pt}, y tick label style={font=\footnotesize},
  ylabel={\% resolved}, label style={font=\footnotesize}, ylabel shift=-2pt,
  legend style={font=\footnotesize, draw=none, fill=none, at={(0.5,1.02)}, anchor=south, legend columns=2},
  nodes near coords, every node near coord/.append style={font=\scriptsize, rotate=90, anchor=west, /pgf/number format/fixed, /pgf/number format/fixed zerofill, /pgf/number format/precision=1}]
\addplot[fill=black!15] coordinates {(gpt6,58.3) (gpt,50.0) (glm,38.9) (hy4,47.2) (ds,44.4) (kimi,44.4)};
\addplot[fill=black!75] coordinates {(gpt6,38.9) (gpt,30.6) (glm,25.0) (hy4,36.1) (ds,30.6) (kimi,30.6)};
\legend{Official, + Probes}
\end{axis}
\end{tikzpicture}
\end{minipage}\hfill
\begin{minipage}[b]{0.47\linewidth}
\centering
\begin{tikzpicture}[baseline=(current axis.south)]
\begin{axis}[ybar=0.5pt, scale only axis, width=0.85\linewidth, height=2.5cm, ymin=0, ymax=90, ytick={0,20,...,80}, bar width=6.5pt,
  xmin=-0.5, xmax=3.8, xtick={0,1,2.3,3.3}, clip=false,
  xticklabels={{GPT-6\\Sol}, {DeepSeek\\V4 Pro}, {GPT-6\\Sol}, {DeepSeek\\V4 Pro}},
  area legend, ymajorgrids, grid style={dashed,gray!30},
  x tick label style={font=\footnotesize, align=center, inner sep=1pt}, y tick label style={font=\footnotesize},
  ylabel={\% resolved}, label style={font=\footnotesize}, ylabel shift=-2pt,
  legend style={font=\footnotesize, draw=none, fill=white, at={(0.98,0.96)}, anchor=north east, legend columns=1},
  nodes near coords, every node near coord/.append style={font=\scriptsize, rotate=90, anchor=west, /pgf/number format/fixed, /pgf/number format/fixed zerofill, /pgf/number format/precision=1}]
\addplot[fill=black!15] coordinates {(0,68.4) (1,52.6) (2.3,5.3) (3.3,5.3)};
\addplot[fill=black!75] coordinates {(0,52.6) (1,42.1) (2.3,0.0) (3.3,0.0)};
\legend{Official, + Probes}
\draw[gray!70] (axis cs:1.65,0) -- (axis cs:1.65,90);
\node[anchor=south, font=\footnotesize, inner sep=2pt] at (axis cs:0.5,90) {Synthesized prompt};
\node[anchor=south, font=\footnotesize, inner sep=2pt] at (axis cs:2.8,90) {Original description};
\end{axis}
\end{tikzpicture}
\end{minipage}
\par
\begin{minipage}[t]{0.48\linewidth}
\caption{Resolved rate (\%) on the \chk{36} tasks, with and without the \probes{}.
}
\label{fig:probe-ablation}
\end{minipage}\hfill
\begin{minipage}[t]{0.50\linewidth}
\caption{Resolved rate (\%) on the \chk{19} Megatron-LM tasks. \emph{Original description} prompts the agent with the PR's own description on GitHub.}
\label{fig:ablation-prompt}
\end{minipage}
\end{figure}

 \textbf{Hidden-probe synthesis.}
The official tests alone overstate how many tasks agents resolve.
We grade every submission twice, on the official tests alone (Set~A) and with the \probes{} added (Set~B; Figure~\ref{fig:probe-ablation}).
Adding the \probes{} lowers every model's resolved rate by \chk{11} to \chk{19} percentage points.
Of the \chk{216} submissions, \chk{102} pass all official tests, but only \chk{69} of them also pass the \probes{}.
The \probes{} thus reject \chk{32}\% of the submissions that the official tests accept.

 \textbf{Prompt synthesis.}
\benchname{} cannot distinguish models without prompt synthesis.
We replace the synthesized prompt with the PR's original description on the Megatron-LM tasks (Figure~\ref{fig:ablation-prompt}).
Once the \probes{} are included, neither GPT-6 Sol nor DeepSeek V4 Pro resolves a single task.
Maintaining comprehensive AI infrastructure is hard, and a prompt that omits the required interfaces makes even capable agents fail for reasons unrelated to the task.
With the synthesized prompt, the resolved rate is at most \chk{52.6}\%; the prompt therefore does not hand the agent the implementation.

\paragraph{Resource-efficient Execution Environment.}

We compare our execution environment against a baseline that builds one Docker image per task, with each Dockerfile following the task's official implementation. \benchname{} reduces storage by \upd{57}\% on Megatron-LM, from \upd{80.7} to \upd{34.8}\,GB, and by \upd{52}\% on vLLM, from \upd{206.0} to \upd{98.8}\,GB. The saving grows with the number of tasks: on Megatron-LM, it rises from \chk{17}\% with two tasks to \chk{57}\% with all \chk{19}, and on vLLM from \chk{27}\% to \chk{52}\%. This suggests great scalability as \benchname{} expands to more tasks. \benchname{} also shortens environment build time by \upd{14}\% on Megatron-LM and \upd{13}\% on vLLM.

\subsection{Failure Analysis}
\label{sec:current-cases}

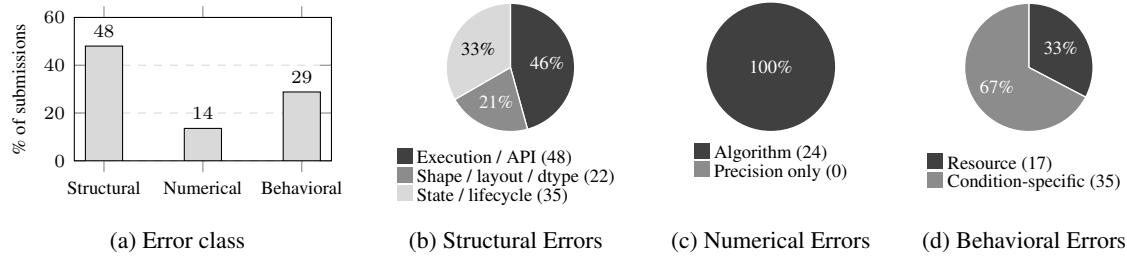
\begin{figure}[t]
\centering
\begin{minipage}[t][3.25cm][s]{0.31\linewidth}
\vspace{0pt}\centering
\begin{tikzpicture}
\begin{axis}[ybar, scale only axis, width=0.72\linewidth, height=1.9cm, ymin=0, ymax=60, bar width=14pt,
  enlarge x limits=0.15, ymajorgrids, grid style={dashed,gray!30}, symbolic x coords={S,N,B}, xtick=data,
  xticklabels={Structural,Numerical,Behavioral}, tick label style={font=\scriptsize}, label style={font=\scriptsize},
  ylabel={\% of submissions}, nodes near coords, every node near coord/.append style={font=\scriptsize,
  /pgf/number format/fixed, /pgf/number format/precision=0}]
\addplot[fill=black!15] coordinates {(S,48.02) (N,13.56) (B,28.81)};
\end{axis}
\end{tikzpicture}
\par\vfill
{\footnotesize (a) Error class}
\end{minipage}\hfill
\begin{minipage}[t][3.25cm][s]{0.24\linewidth}
\vspace{0pt}\centering
\begin{tikzpicture}
\fill[black!75, draw=white, line width=0.5pt] (0,0) -- (90.00:0.85cm) arc[start angle=90.00, end angle=-74.57, radius=0.85cm] -- cycle;
\node[font=\scriptsize, text=white] at (7.71:0.493cm) {46\%};
\fill[black!45, draw=white, line width=0.5pt] (0,0) -- (-74.57:0.85cm) arc[start angle=-74.57, end angle=-150.00, radius=0.85cm] -- cycle;
\node[font=\scriptsize, text=white] at (-112.29:0.493cm) {21\%};
\fill[black!15, draw=white, line width=0.5pt] (0,0) -- (-150.00:0.85cm) arc[start angle=-150.00, end angle=-270.00, radius=0.85cm] -- cycle;
\node[font=\scriptsize, text=black] at (-210.00:0.493cm) {33\%};
\end{tikzpicture}\\[3pt]
{\scriptsize\begin{tabular}{@{}l@{}}\tikz\fill[black!75] (0,0) rectangle (2mm,2mm); Execution / API (48)\\[-1pt]\tikz\fill[black!45] (0,0) rectangle (2mm,2mm); Shape / layout / dtype (22)\\[-1pt]\tikz\fill[black!15] (0,0) rectangle (2mm,2mm); State / lifecycle (35)\end{tabular}}
\par\vfill
{\footnotesize (b) Structural Errors}
\end{minipage}\hfill
\begin{minipage}[t][3.25cm][s]{0.20\linewidth}
\vspace{0pt}\centering
\begin{tikzpicture}
\fill[black!75] (0,0) circle (0.85cm);
\node[font=\scriptsize, text=white] at (0,0) {100\%};
\end{tikzpicture}\\[3pt]
{\scriptsize\begin{tabular}{@{}l@{}}\tikz\fill[black!75] (0,0) rectangle (2mm,2mm); Algorithm (24)\\[-1pt]\tikz\fill[black!45] (0,0) rectangle (2mm,2mm); Precision only (0)\end{tabular}}
\par\vfill
{\footnotesize (c) Numerical Errors}
\end{minipage}\hfill
\begin{minipage}[t][3.25cm][s]{0.22\linewidth}
\vspace{0pt}\centering
\begin{tikzpicture}
\fill[black!75, draw=white, line width=0.5pt] (0,0) -- (90.00:0.85cm) arc[start angle=90.00, end angle=-27.69, radius=0.85cm] -- cycle;
\node[font=\scriptsize, text=white] at (31.15:0.493cm) {33\%};
\fill[black!45, draw=white, line width=0.5pt] (0,0) -- (-27.69:0.85cm) arc[start angle=-27.69, end angle=-270.00, radius=0.85cm] -- cycle;
\node[font=\scriptsize, text=white] at (-148.85:0.493cm) {67\%};
\end{tikzpicture}\\[3pt]
{\scriptsize\begin{tabular}{@{}l@{}}\tikz\fill[black!75] (0,0) rectangle (2mm,2mm); Resource (17)\\[-1pt]\tikz\fill[black!45] (0,0) rectangle (2mm,2mm); Condition-specific (35)\end{tabular}}
\par\vfill
{\footnotesize (d) Behavioral Errors}
\end{minipage}
\caption{(a) Share of the graded submissions with at least one error of each class. (b)--(d) Error types within each class; counts are submissions with at least one failing case of the type, and a submission can count under several types.}
\label{fig:case-taxonomy}
\end{figure}

\textbf{Corpus and taxonomy.}
We analyze the \chk{1{,}754} failing test cases of the graded submissions of the five models other than GPT-6 Sol and group most of them into three error classes, from the most basic to the most demanding (Figure~\ref{fig:case-taxonomy}).
The most basic is a \textbf{structural error}, in which the code does not fit the program around it because it fails to execute, breaks an API, mismatches a tensor's shape, layout, or dtype, or mishandles state or lifecycle.
Once the code fits, a \textbf{numerical error} means it computes wrong values, because of a wrong algorithm or lost precision.
Beyond correct values, a \textbf{behavioral error} means the code misses a requirement of the task, such as lowering peak memory or working under CUDA graphs.
A submission counts under every class that appears among its failing test cases.

\begin{figure}[t]
\centering
\begin{minipage}[b]{0.29\linewidth}
\centering
\begin{tikzpicture}[baseline=(current axis.south)]
\begin{axis}[scale only axis, grid style={dashed,gray!30}, tick label style={font=\footnotesize}, label style={font=\footnotesize}, nodes near coords, every node near coord/.append style={font=\scriptsize, /pgf/number format/fixed, /pgf/number format/precision=0}, width=0.74\linewidth, height=2.2cm, ymin=0, ymax=80, bar width=12pt,
  enlarge x limits=0.45, ymajorgrids, symbolic x coords={G,O}, xtick=data, xticklabels={{CUDA-graph\\tasks},{Other\\tasks}},
  x tick label style={font=\scriptsize, align=center}, ylabel={\% of submissions}]
\addplot[ybar, fill=black!55] coordinates {(G,60.00) (O,14.65)};
\end{axis}
\end{tikzpicture}
\end{minipage}\hfill
\begin{minipage}[b]{0.36\linewidth}
\centering
\begin{tikzpicture}[baseline=(current axis.south)]
\begin{axis}[xbar, scale only axis, grid style={dashed,gray!30}, tick label style={font=\footnotesize}, label style={font=\footnotesize}, nodes near coords, every node near coord/.append style={font=\scriptsize, /pgf/number format/fixed zerofill, /pgf/number format/precision=1}, width=0.58\linewidth, height=2.2cm, xmin=0, xmax=80, bar width=5pt,
  enlarge y limits=0.14, xmajorgrids, symbolic y coords={R,E,N,S,F}, ytick=data, yticklabels={Resource,Error path,Numerical,Structure,Functional},
  xlabel={\% of assertions}]
\addplot[fill=black!55] coordinates {(1.81,R) (1.99,E) (12.02,N) (27.86,S) (56.33,F)};
\end{axis}
\end{tikzpicture}
\end{minipage}\hfill
\begin{minipage}[b]{0.33\linewidth}
\centering
\begin{tikzpicture}[baseline=(current axis.south)]
\begin{axis}[scale only axis, width=0.65\linewidth, height=2.2cm, ymin=0, ymax=80,
  xmin=-0.45, xmax=1.45, xtick={0,1},
  xticklabels={{GPT-5.6 Sol}, {GPT-5.6 Sol\\+ our skill}},
  ymajorgrids, grid style={dashed,gray!30},
  x tick label style={font=\scriptsize, align=center},
  y tick label style={font=\footnotesize},
  ylabel={\% resolved}, label style={font=\footnotesize},
  clip=false,
  nodes near coords, every node near coord/.append style={font=\scriptsize, /pgf/number format/fixed, /pgf/number format/fixed zerofill, /pgf/number format/precision=1}]
\addplot[ybar, fill=black!15, draw=black!40, bar width=12pt] coordinates {(0,42.1)};
\addplot[ybar, fill=black!75, bar width=12pt] coordinates {(1,52.6)};
\draw[->, thick, black!55] (axis cs:0.28,50) -- (axis cs:0.72,58);
\node[font=\scriptsize, inner sep=1pt] at (axis cs:0.5,70) {$+10.5\%$};
\end{axis}
\end{tikzpicture}
\end{minipage}
\par
\begin{minipage}[t]{0.29\linewidth}
\caption{\raggedright Share of submissions with a condition-specific error in different tasks.}
\label{fig:case-cudagraph}
\end{minipage}\hfill
\begin{minipage}[t]{0.36\linewidth}
\caption{\raggedright What the \chk{5,532} assertions in agents' own tests check, over \chk{94} traced runs.}
\label{fig:case-assertions}
\end{minipage}\hfill
\begin{minipage}[t]{0.33\linewidth}
\caption{\raggedright Resolved rate (\%) of GPT-5.6 Sol on Megatron-LM tasks, with and without the agent skill.}
\label{fig:skill-ablation}
\end{minipage}
\end{figure}

\textbf{Structural errors dominate, and state errors often escape official tests.}
Structural errors are the most common class, found in \chk{48}\% of the submissions (Figure~\ref{fig:case-taxonomy}a).
Within the class, execution and API errors lead, followed by state and lifecycle errors (Figure~\ref{fig:case-taxonomy}b).
State errors often escape the official tests, which check the result of a single call rather than the state that the call leaves behind.
In \chk{15} of the \chk{35} submissions with a state error, only our generated \probes{} catch it.
Together, these errors show that models struggle most with changes that span many files of a large repository, where the code must fit the interfaces of many modules and manage state over a long lifetime.

\textbf{Numerical errors come from wrong math, not lost precision.}
All \chk{24} submissions with a numerical error contain an algorithm error, and \chk{none} fails on precision alone (Figure~\ref{fig:case-taxonomy}c).
The errors are far too large to be rounding noise, so they must come from wrong math rather than from precision lost to differences in hardware or implementation details.
On Megatron-LM PR~2440, for example, one patch produces an indexer loss of \chk{19.95} where the loss must be \chk{0.0}.
Contrary to our own experience as infrastructure developers, where getting the algorithm right is relatively easy and matching numerical precision is the hard part, agents struggle with the algorithm rather than the precision.

\textbf{Behavior breaks under execution modes such as CUDA graphs.}
Of the \chk{51} submissions with a behavioral error, \chk{35} contain an error that appears only under a specific execution mode (Figure~\ref{fig:case-taxonomy}d).
As Figure~\ref{fig:case-cudagraph} shows, \chk{60}\% of the submissions on the \chk{four} tasks built around CUDA graphs have such an error, compared with \chk{15}\% on the other tasks.
On the CUDA-graph tasks, \chk{every} such error stays hidden in the default eager mode and appears only once CUDA graphs are turned on.
Agents appear to write code for eager mode and treat CUDA graphs as a simple on/off switch, overlooking that a CUDA graph replays recorded GPU work without rerunning the surrounding Python code, so any logic that must run or change on every call is silently skipped.

\textbf{Agents rarely test resource requirements like memory, time, or traffic.}
Agents write their own tests in \chk{94} of the \chk{103} traced runs, but fewer than \chk{2}\% of the \chk{5{,}532} assertions check resource usage (Figure~\ref{fig:case-assertions}).
Most assertions check functional results (\chk{56}\%) or structure (\chk{28}\%).
Of the \chk{11} traced runs that fail a resource check, only \chk{one} writes any resource assertion.
Agents check that their code runs and returns the right output, but rarely check whether it delivers the efficiency requirement that the task was written for.

\subsection{Improving Agents with Reusable Skills}
\label{sec:skill}

Based on our analysis in Section~\ref{sec:current-cases}, we build an agent skill for AI infrastructure development around three principles.
\begin{itemize}[leftmargin=*, itemsep=1pt, topsep=2pt]
\item \textbf{Check every caller and every piece of state the change touches.} Structural errors are the most common class. The skill asks the agent to find each caller of a function or class it modifies, keep the signatures, tensor shapes, and dtypes those callers expect, and test stateful objects such as caches and counters over several calls rather than one.
\item \textbf{Compare every new computation with a naive reference.} Numerical errors all come from wrong algorithms. The skill asks the agent to write the formula directly in a few lines of plain PyTorch, compare its code with this reference on small inputs, and check invariants that follow from the math, such as a loss that must be zero or gradients that must be averaged rather than summed across data-parallel ranks.
\item \textbf{Test resource requirements and CUDA-graph behavior.} Behavioral errors come from resource requirements and execution modes that agents rarely test. The skill asks the agent to measure peak memory, runtime, or bytes sent before and after its change whenever the task sets a resource requirement, and to rerun its tests with CUDA graphs enabled and check that the outputs match eager mode.
\end{itemize}
The skill contains no task-specific answers or hidden tests.
On Megatron-LM tasks, it raises the resolved rate of GPT-5.6 Sol from 42.1\% to 52.6\%, a gain of 10.5 \% points (Figure~\ref{fig:skill-ablation}).
This suggests that the failure modes exposed by \benchname{} can guide agent improvements. However, the remaining gap is still large: closing it will also require advances in training methods, model architectures, and data quality, which \benchname{} provides a way to measure.

\section{Conclusion and Future Work}
\label{sec:conclusion}

We asked whether coding agents can develop real-world AI infrastructure, and how we can tell.
The answer to the first is not yet: the best frontier agent resolves fewer than \chk{two in five} tasks, and most of its failures are shared by every other agent.
For the second, a pull request's own tests and description are not a reliable judge; grading becomes reliable only when hidden tests and the prompt are written together. The failures are informative. Agents most often write code that does not fit the program around it; when they compute wrong values, the algorithm rather than precision is at fault; and they rarely test whether a change delivers the memory, time, or traffic savings it promises.
We propose an agent skill that targets these weaknesses and improve the resolved rate.

Two challenges remain in scaling \benchname{}.
The first is the cost of multi-GPU execution: each agent run occupies an eight-GPU node for up to 2.5 hours, and grading reruns the multi-GPU tests.
Evaluating a single model therefore consumes several hundred GPU-hours.
Simulation may lower this cost, for example by emulating multiple ranks and their collective communication on fewer GPUs or on CPUs.
The second challenge is PR curation, which still relies on domain experts to select every PR by hand.
This manual step limits how quickly the benchmark can grow and follow the evolution of its source projects.
Automatic methods for identifying suitable PRs would remove this bottleneck.
We leave both directions to future work.

\clearpage
\bibliography{iclr2027_conference}
\bibliographystyle{iclr2027_conference}

\clearpage
\appendix
\raggedbottom
\section{Details of Test and Prompt Synthesis}
\label{app:synthesis}

\subsection{Hidden-Probe Synthesis}
\label{app:probes}

\textbf{Probe types.}
A \emph{numerical probe} compares an output such as a loss, a gradient, or a model state with a reference taken from an existing implementation (Megatron-LM PR~3922 compares the cached Mamba state with a pre-PR path), an invariance (in PR~1989, changing the chunk size must not change the output), a value derived by hand (PR~2440 recomputes the sparse attention in float64 with plain loops), or a property that must always hold (a KL divergence of a distribution from itself is zero).
A \emph{behavioral probe} measures whether the promised effect occurs and compares it with a value derived from the task, such as the bytes that activation offloading moves (PR~1913) or the elements each rank receives (PR~3213).

\textbf{Acceptance check.}
Algorithm~\ref{alg:check-probes} gives the check.
Besides the \base{} and the adversarial patches, every probe must reject a \emph{hollow patch}, which keeps the interfaces of the \gold{} but makes every function added by the PR raise an exception, so that no probe only checks that the new functions exist.
The promise that each adversarial patch breaks is recorded by the agent that writes it and is not checked by execution.
A PR whose probes do not pass within a fixed number of rounds is left without probes.

\begin{algorithm}[H]
\caption{\textsc{CheckProbes}, the acceptance check of Algorithm~\ref{alg:probe-synthesis}. Every run uses the GPU count used for scoring, and a pass on $\Delta_{\mathrm{gold}}$ requires that no case is skipped.}
\label{alg:check-probes}
\begin{algorithmic}[1]
\Require \probes{} $T_{\mathrm{probe}}$ and adversarial patches $\Delta_1, \dots, \Delta_k$; from Algorithm~\ref{alg:probe-synthesis}, the PR $r$, its \gold{} $\Delta_{\mathrm{gold}}$, the \base{} $\mathcal{S}$, and the official tests $T_{\mathrm{off}}$
\Ensure $\mathit{true}$ if and only if $T_{\mathrm{probe}}$ and $\Delta_1, \dots, \Delta_k$ are accepted
\Function{CheckProbes}{$T_{\mathrm{probe}}, \Delta_1, \dots, \Delta_k$}
  \State $\Delta_{\mathrm{hollow}} \gets \Delta_{\mathrm{gold}}$ with every function added by $r$ raising an exception
  \State $\mathit{ok}_1 \gets$ every probe passes on $\Delta_{\mathrm{gold}}$ in \chk{three} consecutive runs \Comment{accepts the right patch}
  \State $\mathit{ok}_2 \gets$ every probe fails on $\mathcal{S}$, $\Delta_{\mathrm{hollow}}$, and $\Delta_1, \dots, \Delta_k$ \Comment{rejects wrong ones}
  \State $\mathit{ok}_3 \gets$ $\Delta_1, \dots, \Delta_k$ score no worse than $\Delta_{\mathrm{gold}}$ on $T_{\mathrm{off}}$ \Comment{they fool $T_{\mathrm{off}}$}
  \State \Return $\mathit{ok}_1 \wedge \mathit{ok}_2 \wedge \mathit{ok}_3$
\EndFunction
\end{algorithmic}
\end{algorithm}

\section{Task Prompt Template}
\label{app:task-prompt}

This appendix records the unified template used to synthesize every task prompt in Section~\ref{sec:synthesis}.
We show the template with placeholders rather than dumping every instance.
The authoring agent of Section~\ref{sec:synthesis} fills those placeholders from the PR description, the \gold{}, the official tests, the handoff record, and the interface contract.
As in that section, the prompt states \emph{what} the change must do and omits \emph{how} to do it.
The figures below show the resulting template as given to the agent.

\paragraph{Structure.}
Figure~\ref{fig:task-desc} states the specification: what and why the change does, its scope, the behaviors the official tests and \probes{} check, which decisions are left to the agent, and the definition of done.
Figure~\ref{fig:task-instructions} requires a fully autonomous run and forbids looking up the original PR.
Figure~\ref{fig:task-interface} is the interface contract: the names, signatures, shapes, and config knobs that the tests call, with no algorithms.
Figure~\ref{fig:task-environment} states the execution rules for validating on the grading hardware and the final reminder before stopping.

\begin{lstlisting}
# Task: Implement this PR

## PR Title
<ORIGINAL PR TITLE>

## PR Description

> This is a **rewritten** task description, synthesized from the merged change and
> its changed tests. It states the *what*, the *why*, the *scope*, and the
> *definition of done*, while deliberately withholding the algorithms, formulas,
> exact numeric behavior, and precise call signatures -- those are the measured
> part. The exact import/call surface the tests pin is given separately in the
> **Interface contract** block after the instructions.

### 1. TL;DR
**What**: <one-paragraph what -- behavior, not algorithm>
**Why**:  <the motivation / the problem this solves>
**Impact**: <what a user can now do; what stays unchanged when the feature is off>

### 2. Big Picture
**Before**: <data flow / responsibilities before this PR -- no formulas>
**After**:  <what changes in the data flow / new responsibilities -- no formulas>

### 3. Change Scope
| Category | Where | What |
|---|---|---|
| <e.g. New core module> | `<path>` | <what this file does -- no implementation> |
| ... | ... | ... |
| **Tests** | `<changed test path(s)>` | <what the tests cover, at a high level> |

### 4. Behavior the implementation must support
- **<axis name>** -- <observable behavior / constraint that must hold>
- ...
- **<regression axis>** -- existing <X> behavior must remain intact; its regression suite stays green.

### 5. What you must design yourself (the measured part -- NOT specified here)
- <the core algorithm(s) -- scoring, loss, masking, etc.>
- <all numeric behavior the tests assert -- tolerances, scales, exact coincidence conditions -- withheld>
- ...

### 6. Definition of Done
**You are NOT done until you have *executed* the code and watched it load and run.**
Matching the interface contract is necessary but **not sufficient** -- code that
"looks correct" but was never run is a **failed task**. The single most common way
to score **zero** is shipping code that does not import/compile: a trivial syntax
or indentation slip in a shared module (e.g. `<SHARED_FILE>`) breaks `conftest`
and makes the **entire** suite un-collectable -- every test scores 0, even the ones
your code would have passed. Treat the items below as a checklist you must
**actually run**, not merely reason about.

**A. Loads & collects (do this FIRST, re-run after every edit to a shared file).**
   - Every module you created/edited imports cleanly:

         <python -c "import ..."   # one per touched module>

   - The test package still collects with **no** syntax/import errors:

         python -m pytest --collect-only <changed test dir>

   This is the highest-priority gate. If it errors, **stop and fix it first.**

   
   

**B. Reachable / clean no-op.** <the feature is reachable from config/CLI; with it
   off, behavior is unchanged -- adapt to this PR; delete if N/A>

**C. Runs & behaves.** You have executed the behavior: <construct the new objects,
   run forward/backward, check shapes/dtypes>. For any parallel/distributed
   behavior, run **multi-rank** -- a single-rank run cannot exercise it. The
   existing <regression suite> still passes.

**The named changed unit tests are what the grader runs (with <N_RANKS> ranks):**
   - `<changed test path>` -- <one-line of what it covers>
   - `<regression test path>` -- stays green.

**Before you end the session, you MUST paste the output of step A** as evidence
that your code loads and the suite collects. Finishing with a module that fails to
import or a suite that fails to collect is an automatic zero.

---
\end{lstlisting}
\figcaptionhere{Task prompt template, part~1 of~4: specification.}{fig:task-desc}

\begin{lstlisting}
## Instructions

### WARNING: Operating mode: you are running FULLY AUTONOMOUS and UNATTENDED

**There is no human watching this session and no one will answer you.** This is a
headless, automated run. If you ask a question, request confirmation, or hand
control back, **nothing replies -- your run simply ends and the task is graded
as-is, unfinished.** Yielding to a non-existent user is the single easiest way to
fail this task.

Therefore, until the task is complete or you run out of time budget:
- **Never stop to ask** for direction, confirmation, permission, or clarification.
- **Never end your turn with a question** ("Should I...?", "Do you want...?").
- **Do not write a "progress report and hand back."** If you have next steps, **do them.**
- **When uncertain, decide and proceed** -- make the most reasonable assumption
  consistent with the description + contract, act, and verify by running.
- **When stuck on a bug, keep debugging it yourself** (hypothesis -> repro -> re-run).
- **The only acceptable ways to end** are: (a) task complete and validation green,
  or (b) time budget exhausted. Until then, **every turn must end with a tool call.**

You are working on the `<REPO>` repository. The code in the current directory is
the state **before** this PR was merged. Your job is to implement the changes above.

**Do NOT look up the actual PR on GitHub** -- implement it from the description and
the interface contract below.

> The section above tells you *what to build and how it must behave*. The block
> below pins only the *seam the tests touch*: exact module paths, class/function
> names, parameter order/defaults, allowed string values, return shapes, dtypes,
> config field names, submodule field names, and the declared types your code is
> handed. It states no algorithm, formula, numeric value, or threshold. Match every
> name precisely; how you satisfy them is the measured work.
\end{lstlisting}
\figcaptionhere{Task prompt template, part~2 of~4: autonomous run and no lookup of the original PR.}{fig:task-instructions}

\begin{lstlisting}
## Interface contract (the tests import/call these exact names -- match them precisely)

### Module `<module.path>`
Must be importable at top level and expose:
- `<function/class signature>` -- <return shape/dtype, return ORDER, error substring
  for pytest.raises(match=...), which args are keyword-only / have defaults>
- ...

Inputs it receives:
- `<param>: <TypeName>` -- (<pre-existing type | new in this PR>)
- ...

### Config knobs -- `<ConfigClass>` (`<path>`) + matching CLI (`<args path>`)
- `<field: type>` - CLI `<--flag>`
- ...

### What is intentionally NOT specified
<the algorithm/math/tolerances/layout -- restate that these are the measured part.>
\end{lstlisting}
\figcaptionhere{Task prompt template, part~3 of~4: interface contract.}{fig:task-interface}

\begin{lstlisting}
## Environment: you MUST validate your changes on real hardware

You are running **inside the real <REPO> unit-test image** -- the same environment
used to evaluate this PR. The full runtime is installed, **all GPUs on this node
are attached**, and the test data is mounted at `<DATA_MOUNT>`.

`import <repo>` resolves to **this working tree** -- the code you edit is what runs.
Confirm with (should print a path under `<CODE_MOUNT>`):

    python -c "import importlib.util as u; s = u.find_spec('<repo>'); print(list(s.submodule_search_locations or []) or s.origin)"

**Validation is mandatory.** Writing code and reasoning that it matches the
contract is *not enough* -- run it. Do the following **in order**; do not declare
done until step 1 passes:

1. **Compile / collect gate -- cheapest, highest priority. Re-run after every edit
   to a shared file.**

       <python -c "import ..."   # one per touched module>
       python -m pytest --collect-only <changed test dir>

   If any errors, **STOP and fix it before writing more code.**

2. **Runtime sanity (single process):** construct the new objects, run a forward
   and a backward, check shapes/dtypes.

3. **Distributed checks (multi-rank):** parallel behavior can only be exercised
   **multi-rank** -- a single-rank run proves nothing. Run it the way the grader
   will (<N_RANKS> ranks, per-test <TIMEOUT_S>s timeout):

       python -m torch.distributed.run --standalone --nproc_per_node=<N_RANKS> -m pytest --timeout=<TIMEOUT_S> <path>

   
   
4. **Guard the regression:** run `<regression test path>` to confirm you did not
   break <X>.

The PR's own official test files are **NOT** provided -- **you are responsible for
validation**: write your own small checks, and at minimum ensure every edited
module imports and `pytest --collect-only` is clean. Do not fetch or guess the
hidden tests. **Finishing with untested, uncompiled code is a failed task even if
the code looks correct.**

---

## Before you stop -- final reminder (autonomous, unattended run)

There is **no user to reply to you.** Before ending, ask yourself: is the task
actually complete -- target behavior implemented, `pytest --collect-only` clean,
your forward/backward and the relevant tests run and passing? If **not**, and you
still have time budget, **take the next action right now.** Do **not** stop to ask,
do **not** post a "next steps" summary and hand back -- either ends the run
unfinished and graded as-is. Keep iterating (edit -> import/collect -> run tests ->
fix) until it is genuinely done or you are out of time.
\end{lstlisting}
\figcaptionhere{Task prompt template, part~4 of~4: execution rules and final reminder.}{fig:task-environment}

\section{Detailed Performance Tables}
\label{app:performance-details}

Table~\ref{tab:test-suites} gives the resolved rate under the official tests alone and with the \probes{}.
Table~\ref{tab:resolution-by-repo} expands Table~\ref{tab:resolution} per repository, and Table~\ref{tab:token-components} splits each token total into non-cached input, cache reads, output, and reasoning.
\vspace{-0.4em}
\begin{table}[H]
\centering
\small
\setlength{\tabcolsep}{5pt}
\caption{Resolved rate (\%) under the official tests alone and under the official tests plus the supplementary probes. A task is resolved when every test in the suite passes.}
\label{tab:test-suites}
\begin{tabular}{@{}lll@{}}
\toprule
\multicolumn{1}{@{}c}{Model} & \multicolumn{1}{c}{Official tests} & \multicolumn{1}{c@{}}{Official + probes} \\
\midrule
\multicolumn{3}{@{}l}{\textit{Megatron-LM: 19 tasks}} \\
GPT-6 Sol & 68.4 & 52.6 \\
GPT-5.6 Sol & 52.6 & 42.1 \\
GLM 5.3 & 36.8 & 26.3 \\
HY4 preview & 52.6 & 42.1 \\
DeepSeek V4 Pro & 52.6 & 42.1 \\
Kimi K3 & 42.1 & 31.6 \\
\midrule
\multicolumn{3}{@{}l}{\textit{verl: 7 tasks}} \\
GPT-6 Sol & 100.0 & 42.9 \\
GPT-5.6 Sol & 85.7 & 28.6 \\
GLM 5.3 & 71.4 & 42.9 \\
HY4 preview & 85.7 & 57.1 \\
DeepSeek V4 Pro & 71.4 & 28.6 \\
Kimi K3 & 85.7 & 57.1 \\
\midrule
\multicolumn{3}{@{}l}{\textit{vLLM: 10 tasks}} \\
GPT-6 Sol & 10.0 & 10.0 \\
GPT-5.6 Sol & 20.0 & 10.0 \\
GLM 5.3 & 20.0 & 10.0 \\
HY4 preview & 10.0 & 10.0 \\
DeepSeek V4 Pro & 10.0 & 10.0 \\
Kimi K3 & 20.0 & 10.0 \\
\bottomrule
\end{tabular}
\end{table}

\begin{table}[H]
\renewcommand{\chk}[1]{#1}\renewcommand{\upd}[1]{#1}
\centering
\small
\setlength{\tabcolsep}{5pt}
\caption{Table~\ref{tab:resolution} per repository. Steps is the mean number of completed model turns per task.}
\label{tab:resolution-by-repo}
\begin{tabular}{@{}lllllll@{}}
\toprule
\multicolumn{1}{@{}c}{Model} & \multicolumn{1}{c}{\% Resolved} & \multicolumn{1}{c}{\% Passed} & \multicolumn{1}{c}{Time (min)} & \multicolumn{1}{c}{Steps} & \multicolumn{1}{c}{Cost (\$)} & \multicolumn{1}{c@{}}{Tokens (M)} \\
\midrule
\multicolumn{7}{@{}l}{\textit{Megatron-LM: \chk{19} tasks}} \\
GPT-6 Sol & \underline{\textbf{\chk{52.6}}} & \chk{80.6} & \underline{\textbf{\chk{50.4}}} & 227.6 & \chk{6.66} & \textbf{\upd{27.00}} \\
GPT-5.6 Sol & \textbf{\chk{42.1}} & \chk{80.1} & \textbf{\chk{71.6}} & \textbf{163.4} & \chk{16.90} & \underline{\textbf{\upd{22.67}}} \\
GLM 5.3 & \chk{26.3} & \textbf{\chk{84.4}} & \chk{149.5} & 203.2 & \chk{12.79} & \chk{45.44} \\
HY4 preview & \textbf{\chk{42.1}} & \underline{\textbf{\chk{91.0}}} & \chk{137.8} & 199.5 & \underline{\textbf{\chk{2.14}}} & \upd{32.35} \\
DeepSeek V4 Pro & \textbf{\chk{42.1}} & \chk{78.0} & \chk{103.2} & 186.6 & \textbf{\chk{2.23}} & \chk{42.66} \\
Kimi K3 & \chk{31.6} & \upd{77.0} & \upd{148.1} & \underline{\textbf{157.4}} & \upd{13.46} & \upd{31.79} \\
\midrule
\multicolumn{7}{@{}l}{\textit{verl: \chk{7} tasks}} \\
GPT-6 Sol & \textbf{\chk{42.9}} & \textbf{\chk{91.0}} & \underline{\textbf{\chk{30.7}}} & \underline{\textbf{88.1}} & \chk{3.72} & \underline{\textbf{\chk{12.31}}} \\
GPT-5.6 Sol & \chk{28.6} & \chk{83.8} & \chk{52.3} & \textbf{114.6} & \chk{11.23} & \textbf{\chk{15.05}} \\
GLM 5.3 & \textbf{\chk{42.9}} & \chk{78.4} & \chk{101.8} & 155.8 & \chk{9.32} & \chk{32.69} \\
HY4 preview & \underline{\textbf{\chk{57.1}}} & \chk{88.9} & \chk{53.7} & 162.0 & \textbf{\chk{1.43}} & \chk{22.16} \\
DeepSeek V4 Pro & \chk{28.6} & \chk{78.5} & \textbf{\chk{42.6}} & 138.6 & \underline{\textbf{\chk{1.26}}} & \chk{26.04} \\
Kimi K3 & \underline{\textbf{\chk{57.1}}} & \underline{\textbf{\chk{92.1}}} & \chk{93.6} & 160.0 & \chk{10.47} & \chk{28.12} \\
\midrule
\multicolumn{7}{@{}l}{\textit{vLLM: \chk{10} tasks}} \\
GPT-6 Sol & \chk{10.0} & \textbf{\upd{66.8}} & \underline{\textbf{\chk{40.8}}} & \underline{\textbf{150.6}} & \chk{4.07} & \underline{\textbf{\chk{15.18}}} \\
GPT-5.6 Sol & \chk{10.0} & \underline{\textbf{\chk{72.2}}} & \textbf{\chk{93.3}} & 246.7 & \chk{25.38} & \textbf{\chk{36.31}} \\
GLM 5.3 & \chk{10.0} & \chk{61.1} & \chk{134.2} & 260.6 & \chk{18.71} & \chk{64.99} \\
HY4 preview & \chk{10.0} & \chk{64.6} & \chk{125.8} & 214.9 & \textbf{\chk{2.47}} & \chk{37.32} \\
DeepSeek V4 Pro & \chk{10.0} & \chk{64.3} & \chk{112.7} & 215.0 & \underline{\textbf{\chk{1.69}}} & \chk{51.21} \\
Kimi K3 & \chk{10.0} & \chk{62.2} & \chk{144.0} & \textbf{176.1} & \chk{15.62} & \chk{40.80} \\
\bottomrule
\end{tabular}
\end{table}

\begin{table}[H]
\renewcommand{\chk}[1]{#1}\renewcommand{\upd}[1]{#1}
\centering
\small
\setlength{\tabcolsep}{4pt}
\caption{Mean parent-session tokens in millions. Input excludes cached tokens. The four components sum to the token totals in Table~\ref{tab:resolution-by-repo} before rounding and average the same runs.}
\label{tab:token-components}
\begin{tabular}{@{}lllll@{}}
\toprule
\multicolumn{1}{@{}c}{Model} & \multicolumn{1}{c}{Input (M)} & \multicolumn{1}{c}{Cache read (M)} & \multicolumn{1}{c}{Output (M)} & \multicolumn{1}{c@{}}{Reasoning (M)} \\
\midrule
\multicolumn{5}{@{}l}{\textit{Megatron-LM}} \\
GPT-6 Sol & \upd{0.336} & \upd{26.600} & \upd{0.037} & \upd{0.030} \\
GPT-5.6 Sol & \chk{0.768} & \chk{21.832} & \chk{0.042} & \chk{0.029} \\
GLM 5.3 & \upd{0.485} & \upd{44.813} & \upd{0.055} & \upd{0.086} \\
HY4 preview & \upd{0.622} & \upd{31.608} & \chk{0.046} & \upd{0.072} \\
DeepSeek V4 Pro & \chk{1.580} & \chk{40.930} & \chk{0.061} & \chk{0.087} \\
Kimi K3 & \chk{1.037} & \chk{30.645} & \chk{0.050} & \chk{0.056} \\
\midrule
\multicolumn{5}{@{}l}{\textit{verl}} \\
GPT-6 Sol & \upd{0.362} & \upd{11.887} & \upd{0.038} & \upd{0.024} \\
GPT-5.6 Sol & \chk{0.441} & \chk{14.555} & \chk{0.035} & \chk{0.023} \\
GLM 5.3 & \chk{0.280} & \chk{32.285} & \chk{0.046} & \chk{0.074} \\
HY4 preview & \chk{0.333} & \chk{21.736} & \chk{0.045} & \chk{0.050} \\
DeepSeek V4 Pro & \chk{1.031} & \chk{24.889} & \chk{0.047} & \chk{0.069} \\
Kimi K3 & \chk{0.279} & \chk{27.745} & \chk{0.055} & \chk{0.045} \\
\midrule
\multicolumn{5}{@{}l}{\textit{vLLM}} \\
GPT-6 Sol & \upd{0.255} & \upd{14.870} & \upd{0.033} & \upd{0.026} \\
GPT-5.6 Sol & \chk{0.948} & \chk{35.262} & \chk{0.056} & \chk{0.044} \\
GLM 5.3 & \chk{1.249} & \chk{63.574} & \chk{0.066} & \chk{0.099} \\
HY4 preview & \chk{0.772} & \chk{36.428} & \chk{0.036} & \chk{0.083} \\
DeepSeek V4 Pro & \chk{0.863} & \chk{50.189} & \chk{0.067} & \chk{0.089} \\
Kimi K3 & \chk{0.671} & \chk{40.000} & \chk{0.054} & \chk{0.077} \\
\bottomrule
\end{tabular}
\end{table}

\section{Failure Analysis: Error Types and Labelling}
\label{app:case-error-audit}

\paragraph{Failing test cases.}
The analysis covers the $36 \times 5 = 180$ submissions of the five models other than GPT-6 Sol.
A \emph{failing test case} is a test case, from the official tests or the \probes{}, that the \gold{} passes and the agent's output patch does not.
We analyze the 1,754 failing test cases and group them by task, test file, error signature, and label each group by reading the failing assertion against the prompt and the submission code.
Table~\ref{tab:case-taxonomy} counts the submissions with at least one case of each type.

\paragraph{Structural errors.}
The code does not fit the program around it.
An \emph{execution or API} error stops the code before any value can be checked: an import or collection failure, a missing symbol, a signature that does not match its callers, or a crash.
A \emph{shape, layout, or dtype} error produces a tensor whose shape, memory layout, or data type differs from what the surrounding code expects.
A \emph{state or lifecycle} error mishandles data that persists across calls, such as caches, buffers, counters, and the ownership of memory, or mishandles the capture and replay of CUDA graphs.

\paragraph{Behavioral errors.}
The code computes correct values but misses a requirement of the task.
A \emph{resource} error misses a stated requirement on memory, speed, communication traffic, or liveness, such as a peak memory that does not drop or a weight sync that sends more entries than have changed.
A \emph{condition-specific} error appears only under a particular condition: a flag, a parallel or CUDA-graph mode, a boundary input, a fallback, or an error path.
A numerical mismatch that appears only under such a condition is labelled condition-specific rather than numerical.

\paragraph{Numerical errors.}
The code runs and fits the program but computes wrong values.
An \emph{algorithm} error implements the wrong computation, such as summing gradients across data-parallel ranks instead of averaging them.
A \emph{precision-only} error implements the right computation with insufficient precision; we assign it only when the same patch passes an exact-logic check.

\begin{table}[H]
\centering
\small
\setlength{\tabcolsep}{5pt}
\caption{Submissions exhibiting each error type, counted once per submission. A submission may exhibit several types. An exception or collection failure is structural; otherwise a failure is labelled by the property its assertion checks.}
\label{tab:case-taxonomy}
\begin{tabular}{@{}llllll@{}}
\toprule
\multicolumn{1}{@{}c}{Class} & \multicolumn{1}{c}{Error type} & \multicolumn{1}{c}{Megatron-LM} & \multicolumn{1}{c}{verl} & \multicolumn{1}{c}{vLLM} & \multicolumn{1}{c@{}}{All} \\
\midrule
Structural & Execution / API & 19 & 10 & 19 & 48 \\
Structural & Shape / layout / dtype & 8 & 4 & 10 & 22 \\
Structural & State / lifecycle & 10 & 8 & 17 & 35 \\
Structural & \textbf{Any structural error} & \textbf{30} & \textbf{18} & \textbf{37} & \textbf{85} \\
\midrule
Behavioral & Resource (memory, speed, traffic) & 14 & 1 & 2 & 17 \\
Behavioral & Condition-specific behavior & 9 & 0 & 26 & 35 \\
Behavioral & \textbf{Any behavioral error} & \textbf{22} & \textbf{1} & \textbf{28} & \textbf{51} \\
\midrule
Numerical & Algorithm & 7 & 6 & 11 & 24 \\
Numerical & Precision only & 0 & 0 & 0 & 0 \\
Numerical & \textbf{Any numerical error} & \textbf{7} & \textbf{6} & \textbf{11} & \textbf{24} \\
\bottomrule
\end{tabular}
\end{table}

\end{document}